\documentclass[journal]{IEEEtran}
\ifCLASSINFOpdf
\else
\fi

\usepackage{times} 
\usepackage{xcolor}
\usepackage{soul}
\usepackage[utf8]{inputenc}
\usepackage{url}
\usepackage{amssymb}
\usepackage{epsfig}
\usepackage{graphicx}
\usepackage{amsthm}
\usepackage{amsmath}
\usepackage{booktabs}
\usepackage{subfigure}
\usepackage{algorithm,algorithmic}
\usepackage{color}
\usepackage{caption}
\usepackage{adjustbox}
\usepackage{multirow}
\usepackage{array}
\usepackage{pdfpages}
\usepackage{siunitx}
\usepackage{graphicx}
\usepackage{eucal}

\def\etal{\textit{et al.}~}

\newcommand{\rsec}[1]{Section~\ref{#1}}

\newcommand{\rfig}[1]{Figure~\ref{#1}}
\newcommand{\rtab}[1]{Table~\ref{#1}}
\newcommand{\ralg}[1]{Algorithm~\ref{#1}}

\begin{document}
%
\title{Unified Multi-View Orthonormal Non-Negative Graph Based Clustering Framework}
%
%
%
\author{Liangchen~Liu,
        Qiuhong~Ke, 
        Chaojie~Li,
        Feiping~Nie,
        Yingying~Zhu
\thanks{Liangchen~Liu (liangchen.liu@unimelb.edu.au) is with the University of Melbourne, Melbourne, Australia}
\thanks{Qiuhong~Ke (Qiuhong.Ke@monash.edu) is with the Monash University, Melbourne, Australia}
\thanks{Chaojie Li (Chaojie.li@unsw.edu.au) is with the University of New South Wales, Sydney, Australia}
\thanks{Feiping~Nie (Corresponding author, feipingnie@gmail.com) is with Northwestern Polytechnical University, Xi'an, China}
\thanks{Yingying Zhu (yingying.zhu@uta.edu) is with University of Texas at Arlington, Arlington, TX, USA}
}

\markboth{Journal of \LaTeX\ Class Files,~Vol.~14, No.~8, August~2015}%
{Shell \MakeLowercase{\textit{et al.}}: Bare Demo of IEEEtran.cls for IEEE Journals}
%



\maketitle

\begin{abstract}
Spectral clustering is an effective methodology for unsupervised learning. 
Most traditional spectral clustering algorithms involve a separate two-step procedure and apply the transformed new representations for the final clustering results. 
Recently, much progress has been made to utilize the non-negative feature property in real-world data and to jointly learn the representation and clustering results.
However, to our knowledge, no previous work considers a unified model that incorporates the important multi-view information with those properties, which severely limits the performance of existing methods.

In this paper, we formulate a novel clustering model, which exploits the non-negative feature property and, more importantly, incorporates the multi-view information into a unified joint learning framework: the unified multi-view orthonormal non-negative graph based clustering framework (Umv-ONGC). 
Then, we derive an effective three-stage iterative solution for the proposed model and provide analytic solutions for the three sub-problems from the three stages.
We also explore, for the first time, the multi-model non-negative graph-based approach to clustering data based on deep features.
Extensive experiments on three benchmark data sets demonstrate the effectiveness of the proposed method.
\end{abstract}


%
\IEEEpeerreviewmaketitle

\section{Introduction}
%
%
%
Graphical representations have been widely applied in many advanced clustering methods due to their capability of capturing various relationships among data samples~\cite{shi2000normalized,hagen1992new,ng2001spectral}.
Such methods typically transform data to an undirected, weighted similarity graph, and then obtain the final clustering results through optimisation solutions based on graph theory. 
For example, Shi~\etal~\cite{shi2000normalized} proposed normalised cut; Hagen~\etal~\cite{hagen1992new} proposed ratio cut for data partitioning; Ng~\etal~\cite{ng2001spectral} proposed a well-known prototype of spectral clustering; and Stoer~\etal~\cite{stoer1997simple} provide additional theoretical analysis on the graph problem; Nie~\etal~\cite{nie2016constrained,6030950} proposed Constrained Laplacian Rank and spectral embedded clustering.
Han~\etal~\cite{HanIJCAI17} proposed the seminal Orthogonal and Nonnegative Graph Reconstruction.
These algorithms have been successfully used in many application such as automatic visual attribute discovery~\cite{liu2016unsupervised,liu2018multi},  video event analysis~\cite{7565615}, feature integration~\cite{cai2011heterogeneous}, image clustering~\cite{yang2010image}, foreground focus~\cite{lee2009foreground}, and dimensionality reduction~\cite{Han2012Sparse}.
Especially, the clustering and graph based algorithms become increasingly critical to the modern web system such as top-$k$ page ranking~\cite{Yang:2019:EEH:3299869.3319886, yang2019homogeneous, Wei:2019:PST:3299869.3319873} and web recommender systems~\cite{DBLP:journals/tkde/ShinKSX18}.
Recently, along with the data explosion in the big data era, numerous novel web applications such as web mining~\cite{Wei:2019:PST:3299869.3319873}, social network analysis~\cite{Qin:2017:GSD:3133956.3134086}, local differential privacy~\cite{wang2019collecting}, and link prediction~\cite{fogaras2005scaling} are in urgent need of advanced methods and models to handle the huge amount of new emerging data. 
The graph based algorithms~\cite{XU2019186, 8798692} and fast advancing deep learning techniques provide the promising route for theoretical study and practical usage.

Most of the graph-based clustering methods involve a two-step separating procedure which includes: 1) creating a similarity graph from the data, and then applying an optimisation approach on the resulting graph to compute the new data representation, and 2) obtaining the final clustering results with a simple light-weight clustering algorithm, such as \textit{k}-means or spectral rotation~\cite{stella2003multiclass}, on the new representation.
Great strides have been made on improving the performance via various ameliorations of the model, especially on the natural characteristics of the new transformed representations for the final clustering procedure. 
One of the most important characteristics is the non-negativity of the transformed representation since negative values are invalid in most real-world applications, such as those involving image pixels, text or word frequency. The seminal work by 
Han~\etal~\cite{HanIJCAI17} to preserve the non-negative property in the new transformed representation captures more manifold structure from high dimensional space, which may benefits greatly the clustering task.
However, the critical multi-view information has not been considered in this method. Further, a more general unified framework is required to incorporate more beneficial functionalities such as out-of-sample clustering for unseen data.

To that end, in this paper, we propose a unified multi-view orthonormal Non-negative graph based clustering framework (Umv-ONGC), which is a generalisation of the Orthogonal and Nonnegative Graph-based clustering methods~\cite{HanIJCAI17}.
We incorporate the multiple view information and more advanced graph construction methods into Umv-ONGC and discuss the relationship of this work to previous spectral embedding techniques. Furthermore, we propose a novel iterative technique to optimise the proposed model.


The main contributions of this work include:
\begin{enumerate}

\item We formulate the problem of unified multi-view non-negative framework with orthonormal constraint for clustering. 

\item We propose an iterative algorithm with three separate stages to solve the above problem.

\item We theoretically analyse the connection between the proposed method and previous works~\cite{6030950}, as well as how to apply the modified iterative method to the problems in those methods.

\item We apply the proposed method Umv-ONGC on the deep features extracted from divergent networks to improve clustering performance, which has not been done in previous methods to our knowledge.


\item We conduct extensive experiments to evaluate Umv-ONGC, and the results confirm its effectiveness.
\end{enumerate}

The rest of this paper is organised as follows. 
Notations and background are provided in~\rsec{sec:background}. 
We introduce our unified multi-view orthonormal Non-negative graph based clustering framework, and the iterative procedure to solve the problem in~\rsec{sec:ONGC}.
Discussion of the connection between the proposed work and some previous works is provided in~\rsec{sec:connection}.
Experimental results and analysis are presented in~\rsec{sec:experiments}.
We conclude our work and discuss future directions in~\rsec{sec:conclusion}.


\section{Notations and background}
\label{sec:background}
\textbf{Notations:} In this paper, we use the bold upper-case font to represent matrices such as $\textbf{F}$.
The $i$-th row and $ij$-th element in $\textbf{F}$ are respectively denoted as $\textbf{F}_i$ and $F_{ij}$. 
 $\operatorname{Tr(\cdot)}$ denotes computing the trace of a matrix.
The $\left\| \cdot \right\|_F$ denotes the Frobenius norm of a matrix.
The $\ell_0$, $\ell_1$, and $\ell_2$-norms of matrix are respectively denoted as $\left\| \cdot \right\|_0$, $\left\| \cdot \right\|_1$, and  $\left\| \cdot \right\|_2$.

\subsection{Spectral clustering}
Generally, traditional spectral clustering methods have two main steps~\cite{von2007tutorial}. 
The first step is formed via:
\begin{equation}
    \mathop {\max }\limits_{{\textbf{F}^\top}\textbf{F} = \textbf{I}} \operatorname{Tr}({\textbf{F}^\top}{\textbf{D}^{ - \frac{1}{2}}}\textbf{A}{\textbf{D}^{ - \frac{1}{2}}}\textbf{F}),
    \label{eq:embedding}
\end{equation}
where $\textbf{A}\in\mathbb{R}^{n \times n}$, the initial similarity matrix, is computed from an arbitrary graph construction method.
In the similarity matrix $\textbf{A}$, the elements indicate the connections among samples.
For instance, the similarity between a pair of samples $\textbf{x}_i$ and $\textbf{x}_j$ is represented as the element $A_{ij}$ in $\textbf{A}$.
In \eqref{eq:embedding}, the matrix $\textbf{D}\in\mathbb{R}^{n \times n}$ means the degree matrix of a graph, which is a diagonal matrix. 
The $i$-th diagonal element of $\textbf{D}$ is $\sum_j {A_{ij}}$.
The new transformed representation for each data point resides in an $m$-dimensional space and is denoted as $\textbf{F}\in\mathbb{R}^{n \times m}$.

The second step can be formed as follows:
\begin{equation}
    \mathop {\min }\limits_{\textbf{Y}\in \mathbb{T},\textbf{C}} \left\| {\textbf{F} - \textbf{Y}{\textbf{C}^\top}} \right\|_F^2,
    \label{eq:kmeans}
\end{equation}
where $\textbf{Y}\in \mathbb{T}$ represents the indicator matrix indicating the cluster belonging of each sample. 
In particular, the symbol $\mathbb{T}$ is a set in which each element satisfies a constraint such as $\textbf{H}\in \{{0,1}\}^{n \times c}$, and each row in $\textbf{H}$ holds $||\textbf{H}_{i \cdot}||_0 = 1 $.
%
The matrix $\textbf{C} \in \mathbb{R}^{m \times c}$ denotes the cluster centroid matrix which is constituted by the centre of each cluster.
The number of clusters is denoted as $c$.
As a convention, in many spectral clustering literature~\cite{von2007tutorial}, $m$ is usually set to be equal to $c$.

\subsection{Deep features}
In this work, since we also explore our proposed Umv-ONGC method on deep features, we briefly introduce the related deep features applied. 
The deep learning methods are based on deep neural networks. One of the most important deep learning architectures, the Convolution Neural Network (CNN)~\cite{ufldl}~\footnote{http://cs231n.github.io/convolutional-networks/}, has been proved effective in various applications of machine learning and computer vision, etc.
This technique extracts the low-level feature from the original image or visual data via various layers of the CNN.
Three of most widely used CNN based networks are respectively: VGG16~\cite{simonyan2014very}, GoogLeNet~\cite{szegedy2015going}, and ResNet~\cite{he2016deep}. 
Each of them has its own advantages and properties. 
Performing fusion of the extracted features from them could benefit some specific tasks such as image clustering or recognition. 

VGG16~\cite{simonyan2014very} is an architecture from the VGG group, Oxford.
In order to construct the deep network, this method improves from AlexNet~\cite{krizhevsky2012imagenet} by replacing large filters with small $3\times3$ one. 
Evidence shows stacked smaller size kernel filters are better than filters with larger kernel size.   
The VGG net achieves the top-5 accuracy of 92.3\% on ImageNet.

In GoogLeNet~\cite{szegedy2015going}, It is considered that most of the activations of neurons are either redundant or unnecessary due to the correlation effect. 
Therefore, the techniques to prune such connections are applied to achieve sparse connections so as to reduce computational costs.

ResNet~\cite{he2016deep} focuses on increasing the depth of the net in order to increase the accuracy. 
To attack the problem of vanishing gradient caused by the increasing number of layers, residual models are designed, which let the signal, required to change the weights, to flow directly from early layers to later layers.
ResNet successfully boosts the performance of deep network structure again from the basis of VGG and AlexNet.


\section{Unified multi-view orthonormal non-negative graph based clustering framework}
\label{sec:ONGC}
To solve the problem of non-negativity preserving in the new representations, we first introduce graph Laplacian matrix $\textbf{L}$, where $\textbf{L} =\textbf{I} - {\textbf{D}^{ - \frac{1}{2}}}\textbf{A}{\textbf{D}^{ - \frac{1}{2}}}$, and transform the graph optimisation model into a more standard form as follow:
\begin{equation}
    \mathop {\min }\limits_{{\textbf{F}^\top}\textbf{F} = \textbf{I}} \operatorname{Tr}({\textbf{F}^\top}\textbf{L}\textbf{F}),
    \label{eq:L}
\end{equation}
By introducing the non-negative constraint, we arrive at the objective for the new problem:
\begin{equation}
    \mathop {\min }\limits_{{\textbf{F}^\top}\textbf{F} = \textbf{I}, \textbf{F} \ge \textbf{0}} \operatorname{Tr}({\textbf{F}^\top}\textbf{L}\textbf{F}),
    \label{eq:O}
\end{equation}
However, this problem in \eqref{eq:O} is an NP-hard problem which is hard to solve.
To address this, Han~\etal~\cite{HanIJCAI17} transform the \eqref{eq:O} into \eqref{eq:W}:
\begin{equation}
    \mathop {\min }\limits_{{\textbf{F}^\top}\textbf{F} = \textbf{I}, \textbf{F} \ge \textbf{0}} \left\| {\textbf{W} - \textbf{F}{\textbf{F}^\top}} \right\|_F^2,
    \label{eq:W}
\end{equation}
where $\textbf{W}$ is the symmetric similarity matrix of the graph with the doubly-stochastic property.
Then they introduce another key relaxing variable $\textbf{G}$ and control the extent of relaxation by introducing an additional relaxing term in \eqref{eq:W} as:
\begin{equation}
    \mathop {\min }\limits_{{\textbf{F}^\top}\textbf{F} = \textbf{I}, \textbf{F} \ge \textbf{0}} \left\| {\textbf{W} - \textbf{F}{\textbf{G}^\top}} \right\|_F^2 + \mu \left\| {\textbf{F} - \textbf{G}} \right\|_F^2,
    \label{eq:G}
\end{equation}
where $\mu$ is a mixing parameter.
This model is solved using an iterative algorithm by Han~\etal~\cite{HanIJCAI17} and is named Orthogonal and Nonnegative Graph Reconstruction (ONGR).

To incorporate the significant multi-view information, the model can be considered into a unified framework.
The unified framework can be formulated as: 
\begin{equation}
    \mathop {\min }\limits_{{\textbf{F}^\top}\textbf{F} = \textbf{I}, \textbf{G} \ge \textbf{0}, {{\alpha ^\top}\textbf{1} = 1,\alpha  \ge 0}} \left\| {\sum\limits_v {{\alpha _v}{\textbf{A}_v}} - \textbf{F}{\textbf{G}^\top}} \right\|_F^2 + \mu \left\| {\textbf{F} - \textbf{G}} \right\|_F^2,
    \label{eq:M}
\end{equation}
where $\alpha = {\left[ {\alpha_1,...,\alpha_v...} \right]^\top} $ is the coefficient for the matrix $A_v$ from graph of each view and the ${\bf{1}} = {\left[ {1,1,...} \right]^\top} \in {\left\{ 1 \right\}^{{n_v}}}$ where ${n_v}$ is the number of views.

From the objective, our framework still keeps the advantages from ONGR~\cite{HanIJCAI17} model, such as variable $\textbf{G}$ keeps a non-negative value; The orthonormal constraint still holds to restrict the solution space; the problem s largely simplified and become tractable. 
More important, the multi-view information is seamlessly integrated.

This modified objective can be solved by our proposed iterative optimisation methods which fix the other variables and only solve the specific one at a time.
As to \eqref{eq:M}, we get three subproblems by fixing $\textbf{F}$, $\textbf{G}$, and $\alpha$ at a time respectively.
The three subproblems can be all solved by off-the-shelf optimisation methods.
The whole iterative algorithm of the Umv-ONGC is divided into three stages described below:

\begin{algorithm}[tb]
\begin{algorithmic}[1]
 \REQUIRE{The similarity matrix of graph: $\textbf{A}_v \in \mathbb{R}^{n \times n}$, number of clusters: $c$, mixing parameter: $\mu$, projected dimension: $m$} \
    \STATE Randomly initialise $\textbf{F} \in \mathbb{R}^{n \times m}$ and $\textbf{G} \in \mathbb{R}^{n \times m}$ which both satisfy $\textbf{F}^\top\textbf{F} = \textbf{I}$, $\textbf{G}^\top\textbf{G} = \textbf{I}$.    
    \REPEAT
    \STATE Fix $\textbf{G}$ and $\alpha$ to update $\textbf{F}$ according to \eqref{eq:Mf}-\eqref{eq:Mf6}
    \STATE Fix $\textbf{F}$ and $\alpha$ to update $\textbf{G}$ according to \eqref{eq:Mf7}-\eqref{eq:Mf13}
    \STATE Fix $\textbf{F}$ and $\textbf{G}$ to update $\alpha$ according to \eqref{eq:alpha}-\eqref{eq:solution}
    \UNTIL {convergence}
   \caption{Summary for Umv-ONGC algorithm for solving~\eqref{eq:M}}
 \label{alg1}
 \end{algorithmic}
\end{algorithm}
\subsection{The first stage: fix G and $\alpha$ to solve F}
In this part, we introduce the first stage of the iterative optimisation which fixes $\textbf{G}$ and $\alpha$, then solves $\textbf{F}$.
When we fix $\textbf{G}$ and $\alpha$, the problem of \eqref{eq:M} is transformed into:
\begin{equation}
    \mathop {\min }\limits_{{\textbf{F}^\top}\textbf{F} = \textbf{I}} \left\| {\sum\limits_v {{\alpha _v}{\textbf{A}_v}} - \textbf{F}{\textbf{G}^\top}} \right\|_F^2 + \mu \left\| {\textbf{F} - \textbf{G}} \right\|_F^2,
    \label{eq:Mf}
\end{equation}
Similar to the methods in~\cite{nie2016A} and removing the constant terms, we have 
\begin{equation}
\mathop {\max }\limits_{\textbf{F}^\top\textbf{F}=\textbf{I}} \operatorname{Tr}\left( {\textbf{F}^\top}\left( {\sum\limits_v {{\alpha _v}{\textbf{A}_v}}\textbf{G}+ \mu\textbf{G}} \right)\right).
    \label{eq:Mf2}
\end{equation}
Set $\textbf{M}=\left( {\sum\limits_v {{\alpha _v}{\textbf{A}_v}}\textbf{G}+ \mu\textbf{G}} \right)$, we arrive at:
\begin{equation}
\mathop {\max }\limits_{{\textbf{F}^\top}\textbf{F} = \textbf{I}} \operatorname{Tr}\left( {{\textbf{F}^\top}\textbf{M}} \right).
    \label{eq:Mf3}
\end{equation}
The full SVD of the Matrix $\textbf{M}$ is $\textbf{M} = \textbf{U}\Sigma \textbf{V}^\top$, then we have 
\begin{equation}
\begin{gathered}
\operatorname{Tr}\left( {{\textbf{F}^\top}\textbf{M}} \right) = \operatorname{Tr}({\textbf{F}^\top}\textbf{U}\Sigma {\textbf{V}^\top}) = \operatorname{Tr}(\Sigma {\textbf{V}^\top}{\textbf{F}^\top}\textbf{U}) 
\\ \Rightarrow  \operatorname{Tr}(\Sigma \textbf{Z}) = \sum\limits_{i = 1}^k {\Sigma _{ii}Z_{ii}} 
\end{gathered},
    \label{eq:Mf4}
\end{equation}
where $\textbf{Z}={\textbf{V}^\top}{\textbf{F}^\top}\textbf{U}$ with $\Sigma _{ii}$ and $Z_{ii}$ being the $(ii)$-th elements of the matrix $\Sigma$ and $\textbf{Z}$ respectively.
Note that $\textbf{Z}{\textbf{Z}^\top} = \textbf{I}$ thus $\left| {{Z_{ii}}} \right| \le 1$. On the other hand, since $\Sigma_{ii} \ge 0$ is a singular value of the matrix $\textbf{M}$. 
Then we have
\begin{equation}
 \operatorname{Tr}\left( {{\textbf{F}^\top}\textbf{M}} \right) = \sum\limits_{i = 1}^k {{Z_{ii}}{\Sigma _{ii}}}  \le \sum\limits_{i = 1}^k {{\Sigma _{ii}}}.
    \label{eq:Mf5}
\end{equation}
The equality holds when ${Z_{ii}} = 1,(1 \le i \le k)$. 
That indicates $\operatorname{Tr}\left( {{\textbf{F}^\top}\textbf{M}} \right)$ reaches the maximum when the matrix $\textbf{Z} = [{\textbf{I}_k},0] \in {\mathbb{R}^{k \times m}}$.
As $\textbf{Z}={\textbf{V}^\top}{\textbf{F}^\top}\textbf{U}$, the optimal solution is achieved for the problem in \eqref{eq:Mf3} as:
\begin{equation}
\textbf{F} = \textbf{U}{\textbf{Z}^\top}{\textbf{V}^\top} = \textbf{U}[{\textbf{I}};0]{\textbf{V}^\top},
    \label{eq:Mf6}
\end{equation}
where $\textbf{I}$ is a $k$ dimensional identity matrix.
Furthermore, through compact SVD of the matrix $\textbf{M}$ as $\textbf{M} = \textbf{U}\textbf{S}{\textbf{V}^\top}$ ($\textbf{U} \in {\mathbb{R}^{m \times k}}, \textbf{S} \in {\mathbb{R}^{k \times k}}, \textbf{V} \in {\mathbb{R}^{k \times k}}$), the solution is simplified as $\textbf{F} = \textbf{U}{\textbf{V}^\top}$.

%
%

\subsection{The second stage: fix F and $\alpha$ to solve G}
At this stage where we fix $\textbf{F}$ and solve $\textbf{G}$, the problem of \eqref{eq:M} is transformed into:
\begin{equation}
    \mathop {\min }\limits_{\textbf{G}\geq0} \left\| {\sum\limits_v {{\alpha _v}{\textbf{A}_v}} - \textbf{F}{\textbf{G}^\top}} \right\|_F^2 + \mu \left\| {\textbf{F} - \textbf{G}} \right\|_F^2,
    \label{eq:Mf7}
\end{equation}
We derive it into
\begin{equation}
\resizebox{\hsize}{!}{$
    \mathop {\min }\limits_{\textbf{G}\geq0} \left(\operatorname{Tr}\left({{\textbf{G}^\top}\textbf{G} - 2{\textbf{G}^\top}\sum\limits_v {{\alpha _v}{\textbf{A}_v}} \textbf{F}}\right) + \mu\operatorname{Tr}\left({{\textbf{G}^\top}\textbf{G} - 2{\textbf{G}^\top}\textbf{F}}\right)\right).
    \label{eq:Mf8}
$}
\end{equation}
Which can be transformed into:
\begin{equation}
    \mathop {\min }\limits_{\textbf{G}\geq0} \left(\operatorname{Tr}(\textbf{G}^\top\textbf{G}) - 2\operatorname{Tr}\left(\textbf{G}^\top \left( {\frac{{\sum\limits_v {{\alpha _v}{\textbf{A}_v}} + \mu \textbf{I}}}{{1 + \mu}}} \right)\textbf{F}\right)\right).
    \label{eq:Mf9}
\end{equation}
We denote $\textbf{P}=\left( {\frac{{\sum\limits_v {{\alpha _v}{\textbf{A}_v}} + \mu \textbf{I}}}{{1 + \mu}}} \right)\textbf{F}$ and we can have:
\begin{equation}
    \mathop {\min }\limits_{\textbf{G}\geq0} \left(\operatorname{Tr}(\textbf{G}^\top\textbf{G}) - 2\operatorname{Tr}(\textbf{G}^\top\textbf{P})\right).
    \label{eq:Mf10}
\end{equation}
After applying Lagrange multiplier method, we have:
\begin{equation}
    \operatorname{L}(\textbf{G},\lambda)=\operatorname{Tr}(\textbf{G}^\top\textbf{G})-2\operatorname{Tr}(\textbf{G}^\top\textbf{P})+\sigma\left\|\textbf{G}\right\|^{2}_{\textbf{F}}
    \label{eq:Mf11}
\end{equation}
where $\sigma$ is the Lagrange multiplier.
This problem can be solved by the conjugate gradient or a close form solution. 
Following the methods in~\cite{nie2016constrained}, we change the problem \eqref{eq:Mf10} into vector form:
\begin{equation}
    \mathop {\min }\limits_{G_{ij}\geq0}\sum_{ij}G_{ij}^{2}-2\sum_{ij}G_{ij}P_{ij} \Rightarrow \min\limits_{G_{ij}\geq0}\|\textbf{G}_i-\textbf{P}_i\|_2^2
    \label{eq:Mf12}
\end{equation}
And \eqref{eq:Mf11} becomes:
\begin{equation}
    \operatorname{L}(\textbf{G}_i,\lambda_i)=\|\textbf{G}_i-\textbf{P}_i\|_2^2-\lambda_i^\top \textbf{G}_i
    \label{eq:Mf13}
\end{equation}
where $\lambda_i\geq \textbf{0}$ are the Lagrange multipliers.
The optimal solution $\hat{\textbf{G}_i}$ should satisfy that the derivative of \eqref{eq:Mf13} w.r.t. $\textbf{G}_i$ is equal to 0, so we have $2\hat{\textbf{G}_i}-2\textbf{P}_i-\lambda_i = \textbf{0}$. 
Then the $j$-th element of $\hat{\textbf{G}_i}$ is $\hat{G_{ij}}-P_{ij}-\lambda_{ij}/2=0$. 
Noting that the $G_{ij}\lambda_{ij}=0$ according to the KKT condition.
Therefore, we can have $\hat{G_{ij}}=(P_{ij})_{+}$.

\subsection{The third stage: fix F and G to solve $\alpha$}
In this stage, we introduce the optimisation of the integrated multi-view information in our model.
With $\textbf{F}$ and $\textbf{G}$ fixed, we have:
\begin{equation}
\mathop {\min }\limits_{{{\alpha ^\top}\textbf{1} = 1,\alpha  \ge 0}} \left\| {\sum\limits_v {{\alpha _v}{\textbf{A}_v}} - \textbf{F}{\textbf{G}^\top}} \right\|_F^2 + \mu \left\| {\textbf{F} - \textbf{G}} \right\|_F^2.
    \label{eq:alpha}
\end{equation}
We can discard the second term since it is constant.
Then, we can have:
\begin{equation}
\mathop {\min }\limits_{{{\alpha ^\top}\textbf{1} = 1,\alpha  \ge 0}} \left\| {\sum\limits_v {{\alpha _v}\hat{\textbf{A}_v}} - \textbf{F}{\textbf{G}^\top}} \right\|_F^2.
    \label{eq:alpha1}
\end{equation}
Let us denote the vector $\textbf{B}_v \in \mathbb{R}^{n^2 \times 1}$ from vectorizing the $\hat{\textbf{A}_v}$, and denote the matrix $\textbf{B}$ with its $v$-th column equal to vector $\textbf{B}_v$, and denote vector $\textbf{c}$ by vectorising $\textbf{F}\textbf{G}^\top$.
We can arrive at:
\begin{equation}
\mathop {\min }\limits_{{{\alpha ^\top}\textbf{1} = 1,\alpha  \ge 0}} \left\| \textbf{B} \alpha - \textbf{c} \right\|_2^2,
    \label{eq:alpha2}
\end{equation}
where ${\bf{\alpha }} = {[{\alpha _1},{\alpha _2},...,{\alpha _{{n_v}}}]^\top} \in {\mathbb{R}^{{n_v} \times 1}}$.
This is a standard quadratic programming (QP) problem, which  can be readily solved by an efficient iterative algorithm in ~\cite{RN286} or an existing convex optimisation problem.
The Lagrangian is:
\begin{equation}
{\mathcal{L}} = \frac{1}{2}{{\bf{\alpha }}^\top}{\textbf{B}^\top}\textbf{B}{\bf{\alpha }} - {{\bf{\alpha }}^\top}{\textbf{B}^\top}\textbf{c} - \gamma \left( {{{\bf{\alpha }}^\top}{\bf{1}} - 1} \right) - {{\bf{\lambda }}^\top}{\bf{\alpha }},
    \label{eq:lagrangian}
\end{equation}
where $\gamma$ is a scalar and $\lambda$ is a Lagrangian coefficient vector, both of which are to be determined.
According to KKT condition, 
\begin{equation}
\begin{aligned}
\frac{{\partial {\mathcal{L}}}}{{\partial {\bf{\alpha }}}} &= {\textbf{B}^\top}\textbf{B}{\bf{\alpha }} - {\textbf{B}^\top}\textbf{c} - \gamma {\bf{1}} - {\bf{\lambda }} = 0, \\ 
\alpha^\top 1 &= 1, \\ 
\gamma & = \frac{{1 - {{\bf{1}}^\top}{{\left( {{\textbf{B}^\top}\textbf{B}} \right)}^{ - 1}}{\textbf{B}^\top}\textbf{c} - {{\bf{1}}^\top}{{\left( {{\textbf{B}^\top}\textbf{B}} \right)}^{ - 1}}{\bf{\lambda }}}}{{{{\bf{1}}^\top}{{\left( {{\textbf{B}^\top}\textbf{B}} \right)}^{ - 1}}{\bf{1}}}}.
\end{aligned}
    \label{eq:kkt}
\end{equation}
For simplicity, we can set:
\begin{equation}
\begin{aligned}
\textbf{N} &= {{\bf{1}}^\top}{\left( {{\textbf{B}^\top}\textbf{B}} \right)^{ - 1}}{\bf{1}} \\
\textbf{V} &= {\textbf{B}^\top}\textbf{c} \\
\textbf{Q} &= {\left( {{\textbf{B}^\top}\textbf{B}} \right)^{ - 1}} \\
\textbf{J} &= \textbf{Q}\textbf{V} + \frac{{\textbf{Q}{\bf{1}}}}{\textbf{N}} - \frac{{\textbf{Q}{\bf{1}}{{\bf{1}}^\top}\textbf{Q}\textbf{V}}}{\textbf{N}} \\
\textbf{K} &= \textbf{Q} - \frac{{\textbf{Q}{\bf{1}}{{\bf{1}}^\top}\textbf{Q}}}{\textbf{N}}.
\end{aligned}
    \label{eq:denote}
\end{equation}
Then, the solution can be simply represented as:
\begin{equation}
{\bf{\alpha }} = \textbf{J} + \textbf{K}{\bf{\lambda }}.
    \label{eq:solution}
\end{equation}
According to the KKT condition, the optimal solution $\star{\alpha}$ to~\eqref{eq:alpha} is $\star{\alpha} = (\textbf{J})_+$.

The whole iterative algorithm for the Unified multi-view orthonormal Non-negative graph based clustering framework is summarised in~\ralg{alg1}.

\section{Connection to prior works}
\label{sec:connection}
The proposed method Umv-ONGC framework is related to various previous works~\cite{6030950,dhillon2004unified,nadler2006diffusion}. 
In this section, we theoretically explore the connection between the Umv-ONGC framework and the recently proposed SEC~\cite{6030950} which unify several important works~\cite{wang2009Clustering,NIPS2007_3176} in its SEC framework. 
Particularly, we will theoretically discuss how the Umv-ONGC method can be adapted to those problems and provide promising solutions.
In the SEC~\cite{6030950} method, the clustering problem can be formed as follow:
\begin{equation}
\begin{aligned}
   \mathop {\min }\limits_{{\bf{F}^ \top}{{\bf{F}} } = {\bf{I}},{\bf{\hat{W}}},{\bf{b}}} & \operatorname{Tr}({{\bf{F}}^ \top {\tilde{\bf{L}}}{{\bf{F}}}}) + \\ & \hat{\gamma} \left( {\left\| {{{\bf{X}}^ \top }{\bf{\hat W}} + {{\bf{1}}_n}{{\bf{b}}^ \top } - {\bf{F}}} \right\|_F^2 + \eta {\mathop{\rm Tr}\nolimits} \left( {{{\bf{\hat{W}}}^ \top }{\bf{\hat{W}}}} \right)} \right)
\end{aligned}
\label{eq:SEC}
\end{equation}
where ${\bf{X}} \in {\mathbb{R}^{d \times n}}$ and  ${\bf{\hat{W}}} \in {\mathbb{R}^{d \times c}}$ are respectively feature representations and weight coefficients. $\mu$ and $\hat{\gamma}$ are two parameters. $\tilde{\bf{L}}$ is the normalized Laplacian matrix.

According to Nie~\etal~\cite{6030950}, the linearity regularization is explicitly applied here on the traditional clustering objective function for purpose of controlling the mismatch between the cluster assignment matrix and the low-dimensional embedding of the data as well as enable out-of-sample clustering.
As mentioned in previous sections, the important non-negative characteristic of the new transformed representation can boost the clustering performance.
We ameliorate the objective function~\eqref{eq:SEC} to impose the non-negative constraint and also include the fusion of multiple graphs encode multi-view information as:
\begin{equation}
\begin{aligned}
   & \mathop {\min }\limits_{{{\bf{F}}^ \top }{\bf{F}} = {\bf{I}},{\bf{F}} \ge 0,{\bf{\hat W}},{\bf{b}},{\alpha^\top}1 = 1,\alpha  \ge 0} {\mathop{\rm Tr}\nolimits} \left( {{{\bf{F}}^ \top }\left( {\textbf{I} - \sum\limits_v {{\alpha _v}{\textbf{A} _v}} } \right){\bf{F}}} \right) + \\& \hat \gamma \left( {\left\| {{{\bf{X}}^ \top }{\bf{\hat W}} + {{\bf{1}}_n}{{\bf{b}}^ \top } - {\bf{F}}} \right\|_F^2 + \eta {\rm{Tr}}\left( {{{{\bf{\hat W}}}^ \top }{\bf{\hat W}}} \right)} \right)
\end{aligned}
\label{eq:SECm1}
\end{equation}
Following the steps of solving SEC, we can obtain:
\begin{equation}
\left\{ \begin{array}{l}
{\bf{b}} = {\textstyle{\frac{1}{n}}}{{\bf{F}}^ \top }{{\bf{1}}_n}\\
{\bf{\hat W}} = {\left( {{\bf{X}}{{\bf{X}}^ \top } + \eta {{\bf{1}}_d}} \right)^{ - 1}}{\bf{XF}}
\end{array} \right.
\label{eq:Wb}
\end{equation}
Substituting $\bf{\hat{W}}$ and $\bf{b}$ into~\eqref{eq:SECm1}, we get:
\begin{equation}
\resizebox{\hsize}{!}{$
\begin{aligned}
&\mathop {\min }\limits_{{{\bf{F}}^ \top }{\bf{F}} = {\bf{I}},{\bf{F}} \ge 0,{\alpha^\top}1 = 1,\alpha  \ge 0} {\mathop{\rm Tr}\nolimits} \left( {{{\bf{F}}^ \top }\left( {\bf{I} - \sum\limits_v {{\alpha _v}{\textbf{A} _v}}  + \hat \gamma {{\bf{L}}_g}} \right){\bf{F}}} \right) \\
\Rightarrow 
&\mathop {\min }\limits_{{{\bf{F}}^ \top }{\bf{F}} = {\bf{I}},{\bf{F}} \ge 0,{\alpha^\top}1 = 1,\alpha  \ge 0} \left\| {\left( {\sum\limits_v {{\alpha _v}{\textbf{A} _v}}  - \hat \gamma {{\bf{L}}_g}} \right) - {\bf{F}}{{\bf{F}}^ \top }} \right\|_F^2
  \end{aligned}
\label{eq:SECm2}
$}
\end{equation}
where ${{\bf{L}}_g} = {\tilde{\bf{H}}} - {{\bf{X}}^ \top }{\left( {{\bf{X}}{{\bf{X}}^ \top } + \eta {{\bf{I}}_d}} \right)^{ - 1}}{\bf{X}}$, and $\tilde{{\bf{H}}} = {{\bf{I}}_n} - \frac{1}{n}{{\bf{1}}_n}{{\bf{1}}_n}^\top$ and $\hat{\gamma}$ is the parameter for the ${\bf{L}}_g$. 

Then, we can easily introduce the additional relaxing variable $\bf{G}$ and multi-view information as in Umv-ONGC and reform the problem as:
\begin{equation}
\resizebox{\hsize}{!}{$
   \mathop {\min }\limits_{{\textbf{F}^\top}\textbf{F} = \textbf{I}, \textbf{G} \ge \textbf{0}, {{\alpha ^\top}\textbf{1} = 1,\alpha  \ge 0}} \left\| \left ( {\sum\limits_v {{\alpha _v}\textbf{A}_v} - \hat{\gamma}\textbf{L}_g} \right ) - \textbf{F}{\textbf{G}^\top} \right\|_F^2 + \mu \left\| {\textbf{F} - \textbf{G}} \right\|_F^2,
\label{eq:SECm3}
$}
\end{equation}
where the matrix $\textbf{A}_v$ can be simply computed from the Laplacian matrix $\textbf{L}$ of each graph respectively.
Fixing $\bf{G}$ and $\alpha$ to solve $\bf{F}$, we can derive~\eqref{eq:SECm3} into:
\begin{equation}
\mathop {\max }\limits_{{\bf{F}^ \top}{{\bf{F}} } = {\bf{I}}}  \operatorname{Tr}\left( {{\bf{F}^ \top\tilde{\bf{M}}}} \right)
\label{eq:SECm4}
\end{equation}
where ${\tilde{\bf{M}}} = \left( \textbf{L}_o + \mu \textbf{I} \right) \textbf{G}$ and $\textbf{L}_o = \left( {\sum\limits_v {{\alpha _v}{\textbf{A}_v}}  - \hat \gamma {{\textbf{L}}_g}} \right)$ is a constant adjacent matrix of a graph.
Similar to \eqref{eq:Mf3}, we compute the new representation $\bf{F}$ via \eqref{eq:Mf3}-\eqref{eq:Mf6}. 

While fixing ${\textbf{F}}$ and $\alpha$ to solve $\textbf{G}$, we can derive~\eqref{eq:SECm3} into:
\begin{equation}
\mathop {\min }\limits_{{\bf{G}} \ge 0} \left( {\operatorname{Tr}\left( {{\bf{G}^ \top}{{\bf{G}}}} \right) - 2\operatorname{Tr}\left( {{\bf{G}}^ \top }{{\tilde{\bf{P}}}} \right)} \right)
\label{eq:SECm5}
\end{equation}
where $\tilde{\bf{P}} = {\frac{{{{\bf{L}}_o} + \mu {\bf{I}}}}{{1 + \mu }}{\bf{F}}}$. 
Then similarly, through~\eqref{eq:Mf10}-\eqref{eq:Mf13}, we can obtain the value of $\bf{G}$ in the current iteration.
Fixing $\textbf{F}$ and $\textbf{G}$ to optimize the $\alpha$, we have \begin{equation}
    \mathop {\min }\limits_{{\alpha ^ \top}\textbf{1} = 1,\alpha  \ge 0} \left\| {\left( {\sum\limits_v {{\alpha _v}{\textbf{A}_v}}  - \hat \gamma {{\bf{L}}_g}} \right) - {\bf{F}}{{\bf{G}}^ \top }} \right\|_F^2 + \mu \left\| {{\bf{F}} - {\bf{G}}} \right\|_F^2.
\end{equation}
We can discard the second term since it is constant:
\begin{equation}
    \mathop {\min }\limits_{{\alpha ^ \top }{\textbf{1}} = 1,\alpha  \ge 0} \left\| {\left( {\sum\limits_v {{\alpha _v}{\textbf{A}_v}}  - \hat \gamma {{\bf{L}}_g}} \right) - {\bf{F}}{{\bf{G}}^ \top }} \right\|_F^2.
\end{equation}
Similar to the derivation from \eqref{eq:alpha1} to \eqref{eq:alpha2}, we can have: 
\begin{equation}
    \mathop {\min }\limits_{{{\alpha ^\top}\textbf{1} = 1,\alpha  \ge 0}} \left\| \textbf{B} \alpha - \textbf{c}_g \right\|_2^2,
    \label{eq:alpha_linpro}
\end{equation}
where the $\textbf{c}_g = \textbf{c} + \textbf{l}_vg$ which includes the vector $c$ from vectorizing the $\textbf{F}\textbf{G}^\top$ and a vector $\textbf{l}_vg$ from vectorizing the $\textbf{L}_g$.
Thus, following the similar derivation steps of \eqref{eq:alpha2}-\eqref{eq:solution}, the optimal $\star\alpha$ can also be derived.
Therefore, the SEC problem with non-negative constrain and multi-view information integration can be solved by the iterative method of Umv-ONGC. 
We summarise the whole iterative procedure in~\ralg{alg2}.
\begin{algorithm}[tb]
\begin{algorithmic}[1]
 \REQUIRE{The similarity matrix of graph: $\textbf{A}_v \in \mathbb{R}^{n \times n}$, number of clusters: $c$, mixing parameters: $\mu$, $\hat{\gamma}$, $\eta$, data matrix $\bf{X}$ and projected dimension: $m$}. \
    \STATE Randomly initialise $\textbf{F} \in \mathbb{R}^{n \times m}$ and $\textbf{G} \in \mathbb{R}^{n \times m}$ which both satisfy $\textbf{F}^\top\textbf{F} = \textbf{I}$, $\textbf{G}^\top\textbf{G} = \textbf{I}$.    
    \REPEAT
    \STATE Fix $\textbf{F}$, $\textbf{G}$, and $\alpha$ to update $\bf{W}$ and $\bf{b}$ according to~\eqref{eq:Wb}
    \STATE Fix $\textbf{G}$, $\alpha$, $\bf{W}$, and $\bf{b}$ to update $\textbf{F}$ according to~\eqref{eq:SECm4}
    \STATE Fix $\textbf{F}$, $\alpha$, $\bf{W}$, and $\bf{b}$ to update $\textbf{G}$ according to \eqref{eq:SECm5}
    \STATE Fix $\textbf{F}$, $\textbf{G}$, $\bf{W}$, and $\bf{b}$ to update $\alpha$ according to \eqref{eq:alpha_linpro}
    \UNTIL {convergence}
   \caption{Summary for the modified iterative Umv-ONGC algorithm for solving~\eqref{eq:SECm1}}
 \label{alg2}
 \end{algorithmic}
\end{algorithm}
Theoretically, our proposed Umv-ONGC framework largely diminishes the solution searching space, while delivering solution, with more preferable characteristic, to the previous SEC problem.

\section{Experiments}
\label{sec:experiments}
Our experiments have four aspects: (1) the evaluation of the proposed novel Umv-ONGC clustering method, (2) the exploration of the method on deep features, (3) convergence analysis, (4) the sensitivity analysis of parameters.
We evaluate the performance of the proposed Umv-ONGC framework on three datasets: Caltech-101~\cite{fei2007learning}, Handwritten dataset~\footnote{https://archive.ics.uci.edu/ml/datasets/Multiple+Features}, and a-Pascal \& a-Yahoo (ApAy) dataset~\cite{Farhadi09describingobjects}.
We first validate our method using traditional hand-crafted features extracted from Caltech-101~\cite{fei2007learning} (two subsets shown in the dataset description part), Handwritten dataset.
Then, we perform a validation on a subset of the Caltech-101 dataset and ApAy dataset especially using three deep features extracted from three different deep networks: VGG16~\cite{simonyan2014very}, GoogLeNet~\cite{szegedy2015going}, and ResNet~\cite{he2016deep}.
The deep features are extracted from applying each deep network on the original sample images in those datasets, respectively.
Those three deep networks with different structures are all pre-trained on ImageNet~\cite{ILSVRC15}.
Two standard metrics are used to evaluate the performance: Clustering Accuracy (ACC) and F1-score.
After that, we perform an empirical convergence analysis of the objective functions of our proposed method to show the efficiency of our optimisation algorithm.
Then, a sensitivity analysis of parameters is provided to show the robustness of the proposed method.

\subsection{Data set descriptions}
Before presenting the experimental results, we first describe the details of four employed benchmark datasets below.
These datasets are summarized in~\rtab{tab:dataset}.
TFC denotes Traditional Feature Concatenated.

\begin{table}[htp]
\centering
\caption{Summary of the dataset descriptions}
\begin{tabular}{ccccc}
\hline
Datasets & Feature type                  & Dimensions & Num of Instances & Classes \\ \hline
\midrule
HW       & TFC & 649        & 2000             & 10      \\
\midrule
Cal7     & TFC & 3768       & 1474             & 7       \\
\midrule
   		 & TFC & 3768       &              &       \\
Cal20    & VGG16                         & 4096       &      2386            &  20    \\
         & GoogLeNet                     & 1024       &                  &         \\
         & ResNet                        & 2048       &                  &         \\
\midrule
	     & VGG16                         & 4096       &                  &       \\
ApAy     & GoogLeNet                     & 1024       &  6340            & 20   \\
         & ResNet                        & 2048       &                  &      \\ \hline
\end{tabular}
\label{tab:dataset}
\end{table}

\textbf{Caltech101~\cite{fei2007learning} --- } is a publicly available dataset.
It has been widely used for various tasks, especially for clustering tasks and object recognition.
There are 8,677 samples of 101 categories. 
According to the previous work~\cite{dueck2007non}, The seven widely used classes are chosen to form a subset denoted as \textbf{Cal7} for evaluation. 
The seven classes, including 1,474 images, are respectively Faces, Motorbikes, Stop-Sign, Windsor-Chair, Dolla-Bill, Garfield, and Snoopy.
Furthermore, we also choose 20 classes to form another testing subset denoted as \textbf{Cal20} with 2,386 images in total.
The 20 classes are respectively Camera, Car-Side, Dollar-Bill, Ferry, Garfield, Hedgehog, Pagoda, Faces, Leopards, Motorbikes, Binocular, Brain, Rhino, Water-Lilly, Windsor-Chair, Wrench, Snoopy, Stapler, Stop-Sign, and Yin-Yang.
Six views/features are extracted from each sample.
We extract various CNN deep features on \textbf{Cal20} subset for deep feature evaluation.

\textbf{HandWritten~(HW)~\cite{Frank2010UCI} ---} dataset consists of handwritten digits from 0 to 9.
This dataset is downloaded from the UCI machine learning repository.
This dataset contains 2,000 samples with a variety of extracted hand-crafted features. 

\textbf{a-Pascal \& a-Yahoo dataset (ApAy)~\cite{Farhadi09describingobjects} --- } comprises two subsets: a-Pascal and a-Yahoo. 
The a-Pascal subset has 20 classes. 
The train set contains 6,340 images and the test set contains 6,355 images.
For the a-Yahoo subset, there are only 12 categories and they are unrelated to the a-Pascal categories. 
As a smaller subset, a-Yahoo contains 2,644 test exemplars. 
The a-Pascal\& a-Yahoo dataset provide rich annotations: 64 attributes are provided for each bounding box of images and 4 hand-crafted features (local texture, HOG, edge and colour descriptor) are provided for each exemplar.
For our purpose of deep feature evaluation, we extract various deep CNN features from this dataset.


\begin{figure*}[htbp!]
    \centering
	\subfigure[]{
		\label{fig:converge:a} 
		\includegraphics[width=0.47\linewidth]{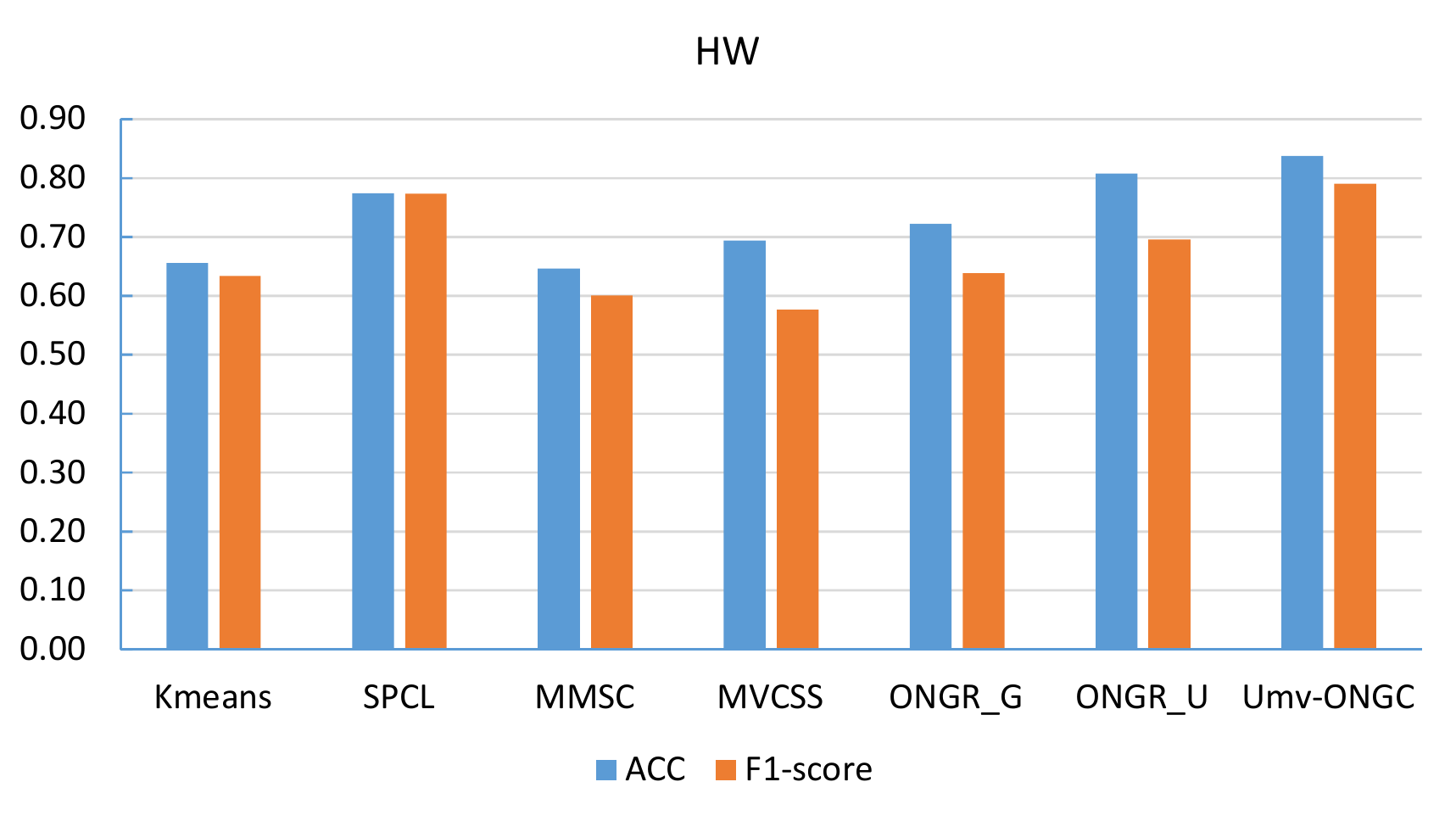}
	}    
	\subfigure[]{
		\label{fig:converge:a} 
		\includegraphics[width=0.47\linewidth]{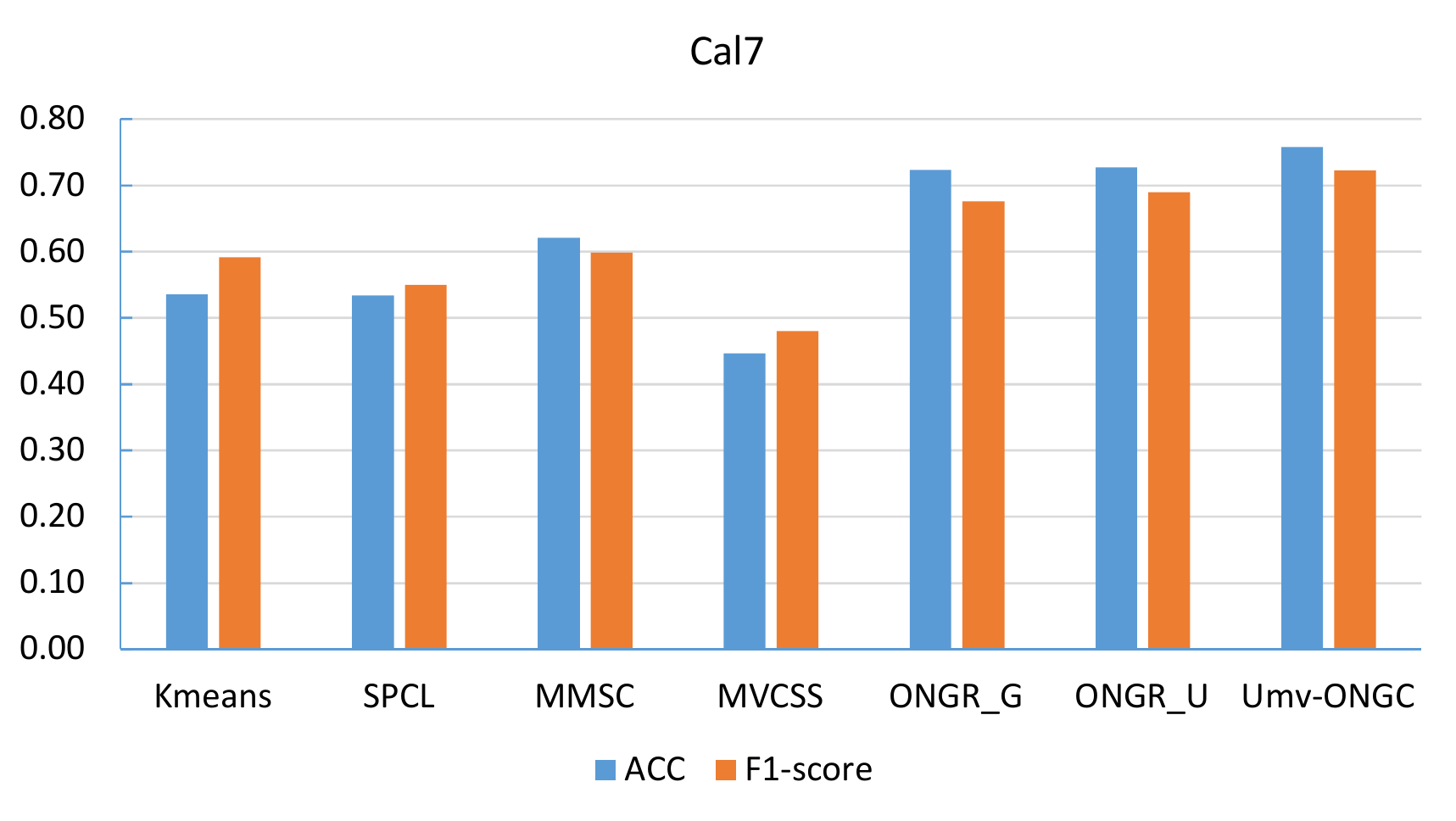}
	}  
	\subfigure[]{
		\label{fig:converge:a} 
		\includegraphics[width=0.47\linewidth]{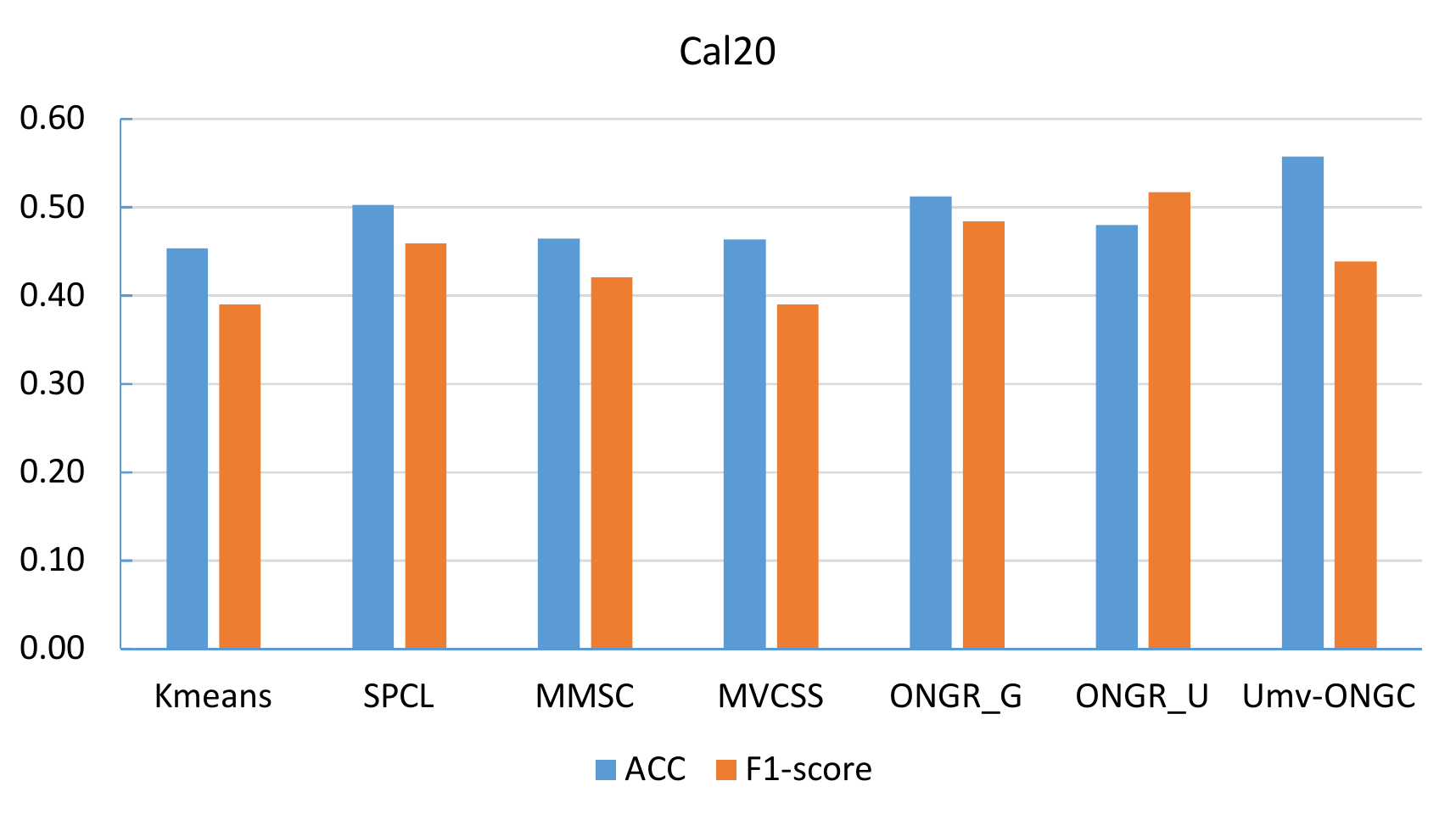}
	}  
	\subfigure[]{
		\label{fig:converge:a} 
		\includegraphics[width=0.47\linewidth]{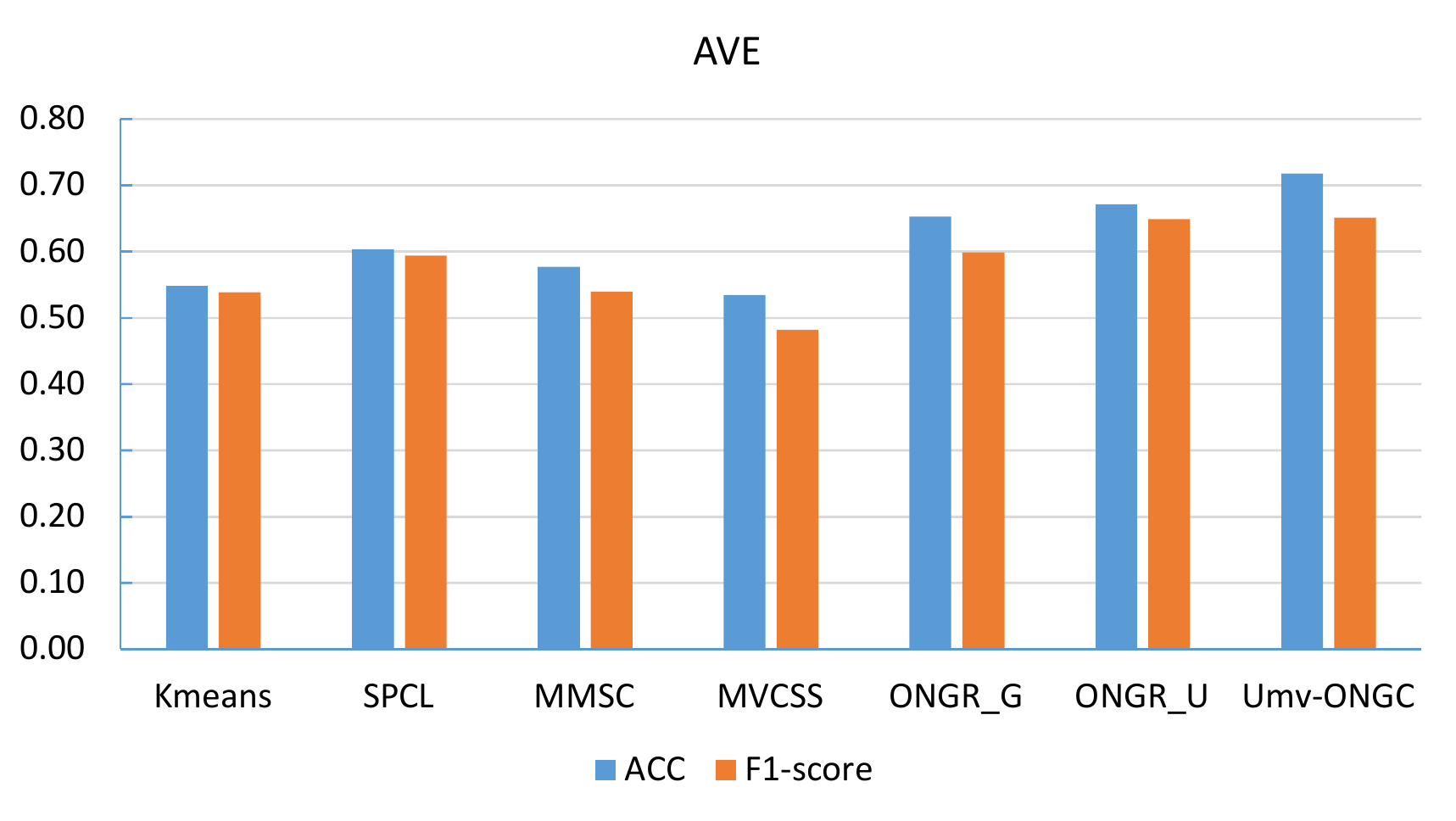}
	}  
	\caption{The clustering performance  on three datasets using traditional hand-crafted features and the average performance (AVE)
	}
	\label{fig:clustering}
\end{figure*}

\begin{table*}[htbp!]
  \centering
  \caption{The clustering performance on deep features evaluated by accuracy}
\begin{tabular}{cc|ccccccc}
\hline
\multicolumn{2}{c|}{ACC}           & Kmeans & SPCL  & MMSC                 & MVCSS                & ONGR\_G & ONGR\_U & Umv-ONGC             \\ \hline
\multirow{4}{*}{Cal20} & VGG16     & 0.521  & 0.564 & \multirow{3}{*}{N/A} & \multirow{3}{*}{N/A} & 0.635   & 0.638   & \multirow{3}{*}{N/A} \\
                       & GoogLeNet & 0.568  & 0.576 &                      &                      & 0.636   & 0.643   &                      \\
                       & ResNet    & 0.532  & 0.525 &                      &                      & 0.646   & 0.636   &                      \\
                       & Fusion    & 0.570  & 0.529 & 0.340                & 0.590                & 0.650   & 0.643   & \textbf{0.865}       \\ \hline
\multirow{4}{*}{ApAy}  & VGG16     & 0.292  & 0.336 & \multirow{3}{*}{N/A} & \multirow{3}{*}{N/A} & 0.382   & 0.361   & \multirow{3}{*}{N/A} \\
                       & GoogLeNet & 0.322  & 0.376 &                      &                      & 0.394   & 0.358   &                      \\
                       & ResNet    & 0.294  & 0.337 &                      &                      & 0.395   & 0.344   &                      \\
                       & Fusion    & 0.309  & 0.350 & 0.394                & 0.306                & 0.398   & 0.378   & \textbf{0.448}       \\ \hline
\end{tabular}
\label{tab:deep_acc}
\end{table*}

\begin{table*}[htbp!]
  \centering
  \caption{The clustering performance on deep features evaluted by F1-score}
\begin{tabular}{cc|ccccccc}
\hline
\multicolumn{2}{c|}{F1-score} & Kmeans & SPCL  & MMSC                  & MVCSS                 & ONGR\_G & ONGR\_U & Umv-ONGC              \\ \hline
                                             & VGG16                         & 0.471  & 0.415 &                       &                       & 0.576   & 0.577   &                       \\
                                             & GoogLeNet                     & 0.544  & 0.388 &                       &                       & 0.629   & 0.552   &                       \\
                                             & ResNet                        & 0.533  & 0.444 & \multirow{-3}{*}{N/A} & \multirow{-3}{*}{N/A} & 0.635   & 0.562   & \multirow{-3}{*}{N/A} \\
\multirow{-4}{*}{Cal20}                      & Fusion                        & 0.536  & 0.485 & 0.286                 & 0.570                 & 0.740   & 0.582   & \textbf{0.853}        \\ \hline
                                             & VGG16                         & 0.221  & 0.242 &                       &                       & 0.301   & 0.238   &                       \\
                                             & GoogLeNet                     & 0.236  & 0.228 &                       &                       & 0.327   & 0.221   &                       \\
                                             & ResNet                        & 0.217  & 0.224 & \multirow{-3}{*}{N/A} & \multirow{-3}{*}{N/A} & 0.325   & 0.220   & \multirow{-3}{*}{N/A} \\
\multirow{-4}{*}{ApAy}                       & Fusion                        & 0.228  & 0.233 & 0.308                 & 0.235                 & 0.333   & 0.304   & \textbf{0.346}        \\ \hline
\end{tabular}
\label{tab:deep_f1}
\end{table*}

\subsection{Experiment setup}
\label{setup}
We introduce the setup of the proposed Umv-ONGC method and the setting of the compared methods.

\textbf{$k$-means} --- is a classic simple and widely used clustering method.
It results in a partitioning of the data space into Voronoi cells.
For this method, we concatenate all traditional hand-crafted features for the evaluation in~\rfig{fig:clustering} and we concatenate all features for the fusion method in~~\rtab{tab:deep_acc} and \rtab{tab:deep_f1}.

\textbf{SPectral CLustering (SPCL)} ---
performs spectral clustering~\cite{ng2001spectral} on a Gaussian graph which is constructed by extracted traditional hand-crafted features and deep features.
This method serves as the prototype of many graph-based clustering approaches. 
The same as in the $k$-means method, we concatenate all traditional hand-crafted features for the evaluation in~\rfig{fig:clustering} and we concatenate all features for the fusion method in~~\rtab{tab:deep_acc} and \rtab{tab:deep_f1}.

\textbf{Multi-View Clustering and Feature Learning via Structured Sparsity (MVCSS)}--- is a multi-view learning model proposed in~\cite{wang2013multi}. 
The effective integration of multiple features is achieved by learning the weight for every feature with respect to each cluster individually following novel joint structured sparsity-inducing norms. 
Following the parameter tuning step in~\cite{wang2013multi}, we choose the best value for our experiments.

\textbf{Multi-Modal Spectral Clustering (MMSC)}~\cite{cai2011heterogeneous} --- is a graph-based multi-modal clustering method that integrates different models by constructing a commonly shared graph Laplacian. 
Following the parameter tuning step in~\cite{cai2011heterogeneous}, we choose the best value for our experiments.

\textbf{The ONGR method} --- two versions are developed, respectively, with the Gaussian graph and the Unsupervised Large Graph Embedding (ULGE) graph~\cite{li2015large}.
Similar to the anchor graph applied in~\cite{HanIJCAI17}, the ULGE graph is a recently proposed more advanced anchor-based graph, which is computationally efficient and can handle large-scale datasets.
The two versions of ONGR are denoted as \textbf{ONGR\_G} (with Gaussian graph) and \textbf{ONGR\_U} (with ULGE graph), respectively.

\textbf{The proposed Umv-ONGR method} --- the important parameters are the mixing parameter $\mu$.
Following~\cite{li2015large}, we tune the mixing parameter $\mu$ 
using the logarithmic scale.
We make the value of $\log_{10} \mu$ ranging from -5 to 5 with an incremental size 1.
For the graph construction, we apply the recently proposed Constrained Laplacian Rank graph CLR ~\cite{nie2016constrained} and the same setting as reported to construct the similarity graph for each view.
We report the average results over five runs for all the experiments.

\begin{figure}[htbp!]
	\centering
	\subfigure[]{
		\label{fig:converge:a} 
		\includegraphics[width=0.8\linewidth]{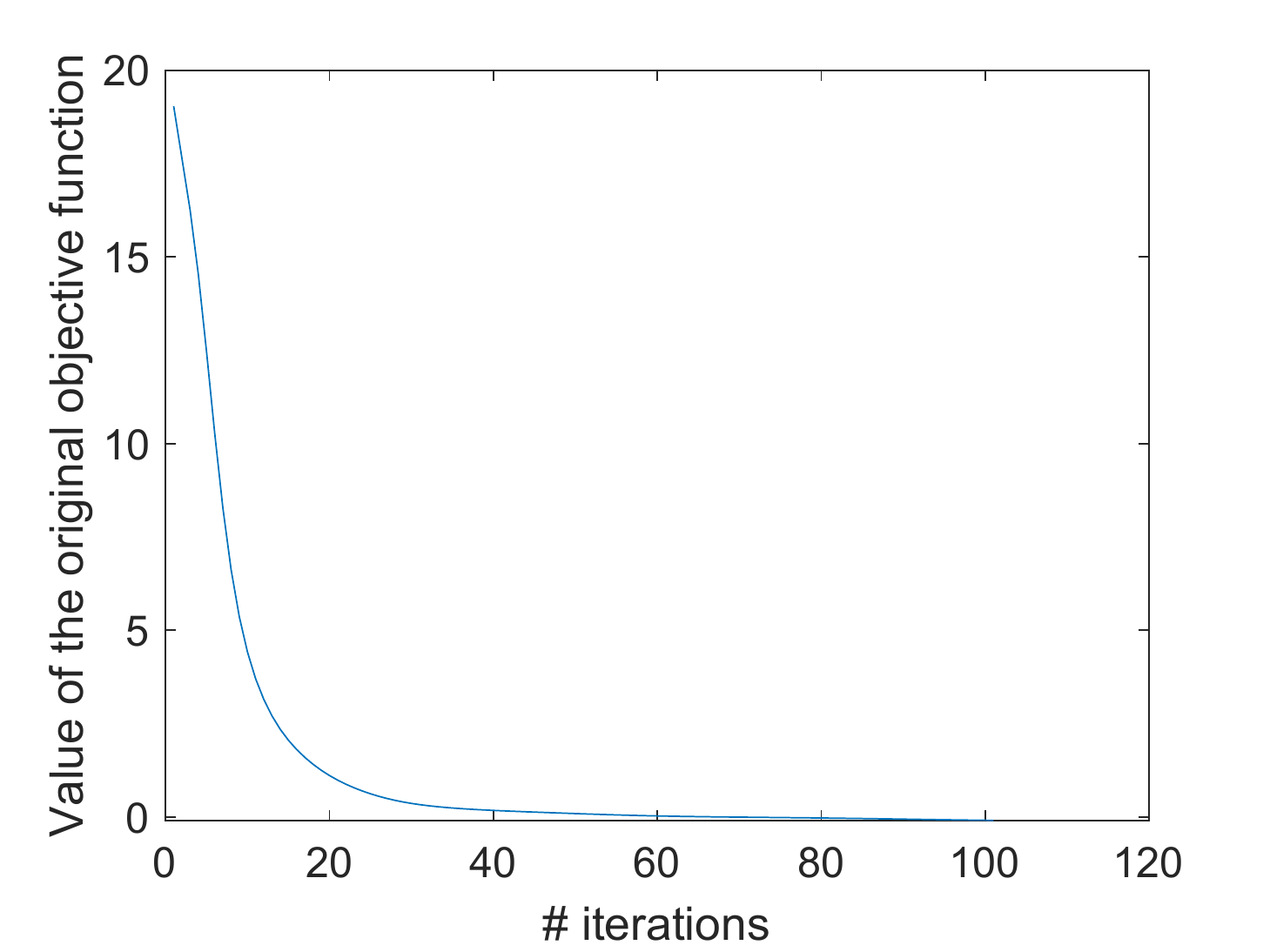}
	}                
	\subfigure[]{
		\label{fig:converge:b} 
		\includegraphics[width=0.8\linewidth]{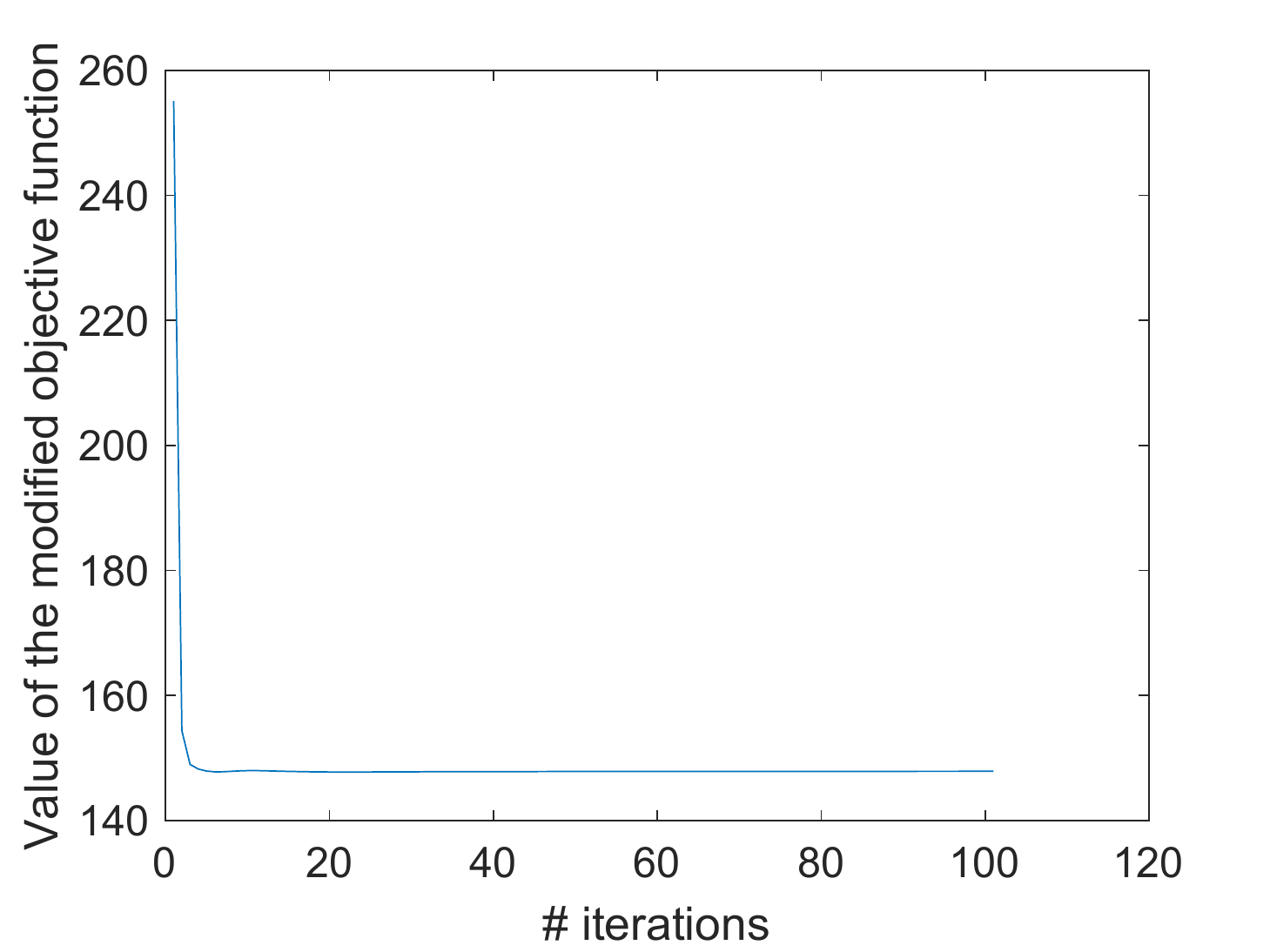}
	}
	\caption{Convergence study. 
		Graph (a) is the convergence curve of original objective function \eqref{eq:O}. 
		Graph (b) is the convergence curve of modified objective function \eqref{eq:M}.
	}
	\label{fig:converge}
\end{figure}

\begin{figure*}[htbp!]
	\centering
	\includegraphics[width=1\linewidth]{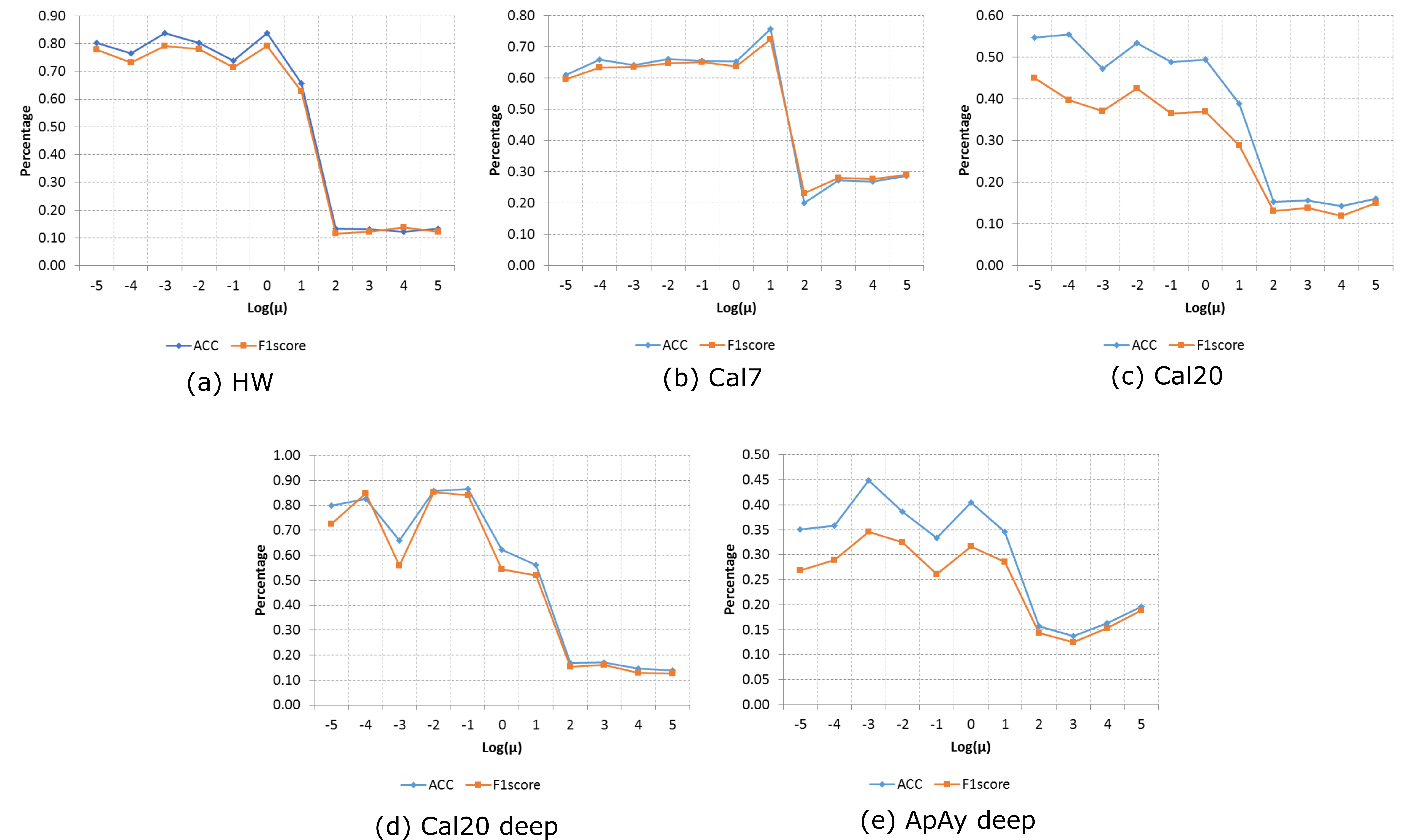}
	\caption{The results of parameter sensitivity evaluation, the first row is the ACC and F1-score from Umv-ONGC with respect to different datasets using traditional features, and the second row using deep features as mixing parameter $\mu$ is varied from -5 to 5 by logarithm to the base 10.
	}
	\label{fig:sen}
\end{figure*}

\subsection{Clustering evaluation}

\rfig{fig:clustering} presents the clustering results of Cal7, Cal20 and HW datasets using traditional features and the average performance.
The results show that the proposed Umv-ONGC method outperforms the comparable clustering methods in most cases.
This validates the multi-view information and novel optimization method are beneficial to the clustering task and boost the performance.
The observation shows that the non-negative and orthonormal constrained models work better than the traditional two-step graph-based methods.
Another observation is that the ONGR\_U version performs better than the ONGR\_G version in some cases.
We argue this is because the chosen number of anchors for this graph is more adequate to capture the data distribution for those datasets.

\subsection{Deep feature evaluation}
In this section, we evaluate the performance of the proposed Umv-ONGC with deep features.
We extract the CNN features before the last full-connected layer.
\rtab{tab:deep_acc} and \rtab{tab:deep_f1} show the results of the proposed method compared with other methods. 
Note that since MMSC, MVCSS and, our Umv-ONGC method are all multi-view clustering methods, we cannot perform the experiments on a single deep feature.
Therefore, only multiple deep feature fusion results are reported here.
From the results, our proposed Umv-ONGC achieves better than or comparable clustering performance to the other methods.
We also note, with appropriate parameter settings, the more advanced multi-view clustering methods MMSC, MVCSS, and our method perform much better than traditional clustering methods $k$-means and SPCL with the simply concatenated feature.
Furthermore, The results from the fusion ones which fuse different CNN features are better than the single CNN feature from each individual deep network. 
This indicates that even for a simple fusion method, the performance still increases. 
Compared to \rfig{fig:clustering}, we can see the CNN deep features performed significantly better than the hand-crafted features and our Umv-ONGC method shows larger performance improvement on deep features than traditional features.
Another observation is that the ONGC\_G performs better than ONGC\_U in most cases. 
This could be because for large datasets and complex features, the chosen number of the anchors of the graph is not enough.
However, the advantage of the ULGE is computational efficiency which benefits large-scale big data processes.
Moreover, the performance of all evaluated methods on the ApAy dataset is not as good as other datasets.
This is probably because there is a large amount of image stretching and compression preprocessing before fitting the data to the network. 
The resolution of samples is various and some of the samples are quite small as they are small parts cropped from whole images.
The pre-processing stage of the deep networks transforms them to the same size. 
From those observations, the current preprocessing for extracting deep features may not be good at dealing with those complicated transformed images.

\subsection{Convergence analysis}

In this part, we empirically analyse the convergence of the proposed Umv-ONGC method in \ralg{alg1}.
We perform this analysis on the Cal20 dataset and set the $\mu$ to the optimal value.
Note that the same convergence property is also observed in other datasets. 
As reported in \rfig{fig:converge}, the results indicate that the proposed methods converge for both original objective~\eqref{eq:O} and modified objective~\eqref{eq:M} within a few iterations. 

Thus, the ONGC method can be efficiently optimised and reasonable results can be achieved within 40 iterations.

\subsection{The sensitivity analysis}
The main parameter in Umv-ONGC is the mixing parameter $\mu$.
It controls the divergence between the new representations $\textbf{F}$ and the relaxing variable $\textbf{G}$ so as to enforce the new representations $\textbf{F}$ to be approximately non-negative.

The results of parameter sensitivity evaluation are shown in \rfig{fig:sen}.
From the results, we can see that within a reasonable range, the proposed method is robust to parameter change such as on HW, Cal7, and Cal20 datasets.
However, some tuning effort is also required for Cal20 deep and ApAy deep datasets; there is a sweet spot in an appropriate range of the mixing parameter $\mu$.
On all datasets, the performance of our Umv-ONGC method suffers a steep drop after the parameter is set too large around and after $10^2$ ($log(\mu)=2$).
  
  
\section{Conclusion and future works}
\label{sec:conclusion}
In this work, we proposed a novel clustering method called unified multi-view orthonormal non-negative graph based clustering framework (Umv-ONGC).
We first formulated a novel unified multi-view non-negative frameworkwith orthonormal constraints for the clustering problem.
Then, we derived a three-stage iterative optimisation algorithm to solve the problem of the model.
The solutions for the subproblems from each stage were also derived.
We also explored, for the first time, the multi-view clustering model for the multiple deep feature clustering.
The experiment results on three benchmark datasets indicated the effectiveness of the proposed method. 
Moreover, we theoretically discussed how the proposed methods connect to previous works and can be simply applied to those problems.

In future work, we will explore our proposed methods for some real-world applications such as web mining, web recommendation system, social network analysis or visual attribute discovery, etc. 
A better deep feature fusion method may also be a good direction to boost the performance under the proposed framework.

\section*{Acknowledgment}

\bibliographystyle{IEEEtran}
\bibliography{egbib}



%

\begin{IEEEbiography}{Liangchen Liu}
received the B.Eng. degrre in information engineering and M.Sc. degree in instrument science and technology from Chongqing University, Chongqing, China, in 2009 and 2012, respectively, and the Ph.D. degree from the University of Queensland, Brisbane, QLD, Australia, in 2017. He is currently a Research Fellow with the University of Melbourne. His current research interests include unsupervised learning, multi-modal learning, detection, segmentation, and visual attribute and its related applications.
\end{IEEEbiography}

\begin{IEEEbiographynophoto}{Qiuhong Ke}
Qiuhong Ke received the Ph.D. degree from The University of Western Australia in 2018. She is currently a Lecturer with the department of Data Science \& AI, Monash University. Her research interests include computer vision and machine learning.
\end{IEEEbiographynophoto}


\begin{IEEEbiographynophoto}{Chaojie Li}
received the B.Eng. degree in electronic science and technology and the M.Eng. degree in computer science from Chongqing University, Chongqing, China, in 2007 and 2011, respectively, and the Ph.D. degree from RMIT University, Melbourne, Australia, in 2017, where he was a Research Fellow for one and a half years. He was a Senior Algorithm Engineer with Alibaba Group. He is a Senior Research Associate at School of Electrical Engineering and Telecommunications, UNSW. He was a recipient of ARC Discovery Early Career Researcher Award in 2020. His current research interests include graph representation learning, distributed optimization and control in smart grid, neural networks, and their application.
\end{IEEEbiographynophoto}

\begin{IEEEbiographynophoto}{Feiping Nie}
(Member, IEEE) received the Ph.D. degree in computer science from Tsinghua University, Beijing, China, in 2009. He is currently a Full Professor with Northwestern Polytechnical University, Xi’an, China. He has published more than 100 papers in the following journals and conferences: IEEE TRANSACTIONS ON PATTERN ANALYSIS AND MACHINE
INTELLIGENCE, INTERNATIONAL JOURNAL OF COMPUTER VISION, IEEE TRANSACTIONS ON IMAGE PROCESSING, IEEE TRANSACTIONS ON NEURAL NETWORKS AND LEARNING SYSTEMS, IEEE TRANSACTIONS ON KNOWLEDGE AND DATA ENGINEERING, ICML, NIPS, KDD, IJCAI, AAAI, ICCV, CVPR, and ACM MM. His papers have been cited more than 20 000 times and the H-index is 78. His research interests are machine learning and its applications, such as pattern recognition, data mining, computer vision, image processing, and information retrieval. Prof. Nie is currently serving as an associate editor or PC member for
several prestigious journals and conferences in the related fields.
\end{IEEEbiographynophoto}


\begin{IEEEbiographynophoto}{Yingying Zhu}
is an Assistant Professor at the Computer Science \& Engineering Department of University of Texas at Arlington (UTA). Dr. Zhu received her Ph.D. in Computer Science from University of Queensland, Australia in 2014. Before she joined UTA, Dr. Zhu was a staff scientist in the Clinical Center, NIH (2019-2020) and a Postdoctoral Scholar at UNC chapel hill (2015-2017) and Cornell University (2017-2019).
\end{IEEEbiographynophoto}




\end{document}